\pdfoutput=1

\documentclass[letterpaper, 10 pt, journal, twoside]{IEEEtran}

\IEEEoverridecommandlockouts                              

\usepackage{multicol}
\usepackage{multirow}
\usepackage{algpseudocode}
\usepackage{color}
\usepackage{xcolor}
\usepackage{graphicx}
\usepackage{bm}
\usepackage{amsmath}
\usepackage{amssymb}
\usepackage{tabularx}
\usepackage{subcaption}
\usepackage{graphicx}
\usepackage{amsmath}
\usepackage{amssymb}
\usepackage{booktabs}
\usepackage{subcaption}
\usepackage{url}
\usepackage{cite}
\usepackage{prettyref}
\usepackage{booktabs}
\usepackage{siunitx}
\usepackage{comment}
\usepackage{booktabs}
\usepackage{hyperref}
\usepackage[capitalize]{cleveref}

\usepackage{tabularx, booktabs}
\usepackage{makecell} 
\usepackage{tabularx} 
\definecolor{Gray}{gray}{0.9}

\usepackage[all]{hypcap}
\usepackage[capitalize]{cleveref}
\crefname{section}{Sec.}{Secs.}
\Crefname{section}{Section}{Sections}
\Crefname{table}{Table}{Tables}
\crefname{table}{Tab.}{Tabs.}

\usepackage[ruled]{algorithm2e} 

\SetAlFnt{\small}
\SetAlCapFnt{\small}
\SetAlCapNameFnt{\small}
\SetAlCapHSkip{0pt}

\newrefformat{fig}{Figure~\ref{#1}}
\newrefformat{sec}{Section~\ref{#1}}
\newrefformat{tab}{Table~\ref{#1}}
\newrefformat{alg}{Algorithm~\ref{#1}}
\newrefformat{eqn}{Equation~\ref{#1}}

\hypersetup{
    colorlinks,
    linkcolor={blue!80!black},
    citecolor={blue!80!black},
    urlcolor={blue!100!black}
}

\makeatletter
\newcommand{\printfnsymbol}[1]{%
  \textsuperscript{\@fnsymbol{#1}}%
}
\makeatother
\newcommand{\sysname}{WeatherDG}

\title{ \sysname:  LLM-assisted Diffusion Model for Procedural Weather Generation in Domain-Generalized Semantic Segmentation
}

\author{
    Chenghao Qian$^*$,
    Yuhu Guo$^*$,
    Yuhong Mo,
    Wenjing Li 
    
    \thanks{C. Qian, and W. Li are with the Transport Studies at Institute at University of Leeds, UK}
    \thanks{Y. Guo and Y. Mo are with Department of Electrical and Computer Engineering, Carnegie Mellon University, USA}
    \thanks{$^*$ denotes equal contribution}
}

\begin{document}
\maketitle

\begin{abstract}

 In this work, we propose a novel approach, namely WeatherDG, that can generate realistic, weather-diverse, and driving-screen images based on the cooperation of two foundation models, i.e, Large Language Model (LLM) and Stable Diffusion (SD). Specifically, we first fine-tune the SD with source data, aligning the content and layout of generated samples with real-world driving scenarios. Then, we propose a procedural prompt generation method based on LLM, which can enrich scenario descriptions and help SD automatically generate more diverse, detailed images. In addition, we introduce a balanced generation strategy, which encourages the SD to generate high-quality objects of tailed classes under various weather conditions, such as riders and motorcycles. This segmentation-model-agnostic method can improve the generalization ability of existing models by additionally adapting them with the generated synthetic data. Experiments on three challenging datasets show that our method can significantly improve the segmentation performance of different state-of-the-art models on target domains. Notably, in the setting of ''Cityscapes to ACDC'', our method improves the baseline HRDA by 13.9\% in mIoU. See the project page for more results:\href{https://jumponthemoon.github.io/WeatherDG.github.io/}{weatherDG.github.io.}

\end{abstract}


\section{Introduction}
\label{sec:intro} 


Semantic segmentation is a fundamental task in autonomous driving. Despite significant achievements in this field, existing models still face serious challenges when deploying in unseen domains due to the well-known domain shift problem \cite{hoyer2022daformer}. In addition, this issue will be more serious when the unseen domains are with adverse weather conditions \cite{4DenoiseNet,Perception}, such as foggy, rainy, snowy, and nighttime scenarios \cite{Lee2021TaskDrivenDI}.

One naive way to solve the above problem is collecting more diverse training data. However, labelling segmentation is a time-consuming process, as we need to annotate every pixel in an image. Hence, domain generalization becomes popular in solving the domain shift problem \cite{jiang2023domain}, in which the goal is to train a model that can generalize to unseen domains using only the given source data. Existing domain generalization methods can be generally divided into two categories: normalization~\cite{peng2022semantic,choi2021robustnet} and data augmentation \cite{ Niemeijer2023DIDEX,Benigmim_2024_CVPR}. In this paper, we focus on the latter category, which is more flexible to different model structures and can be easily integrated with the former techniques.

\begin{figure}[h]
    \centering
    \includegraphics[width=\linewidth]{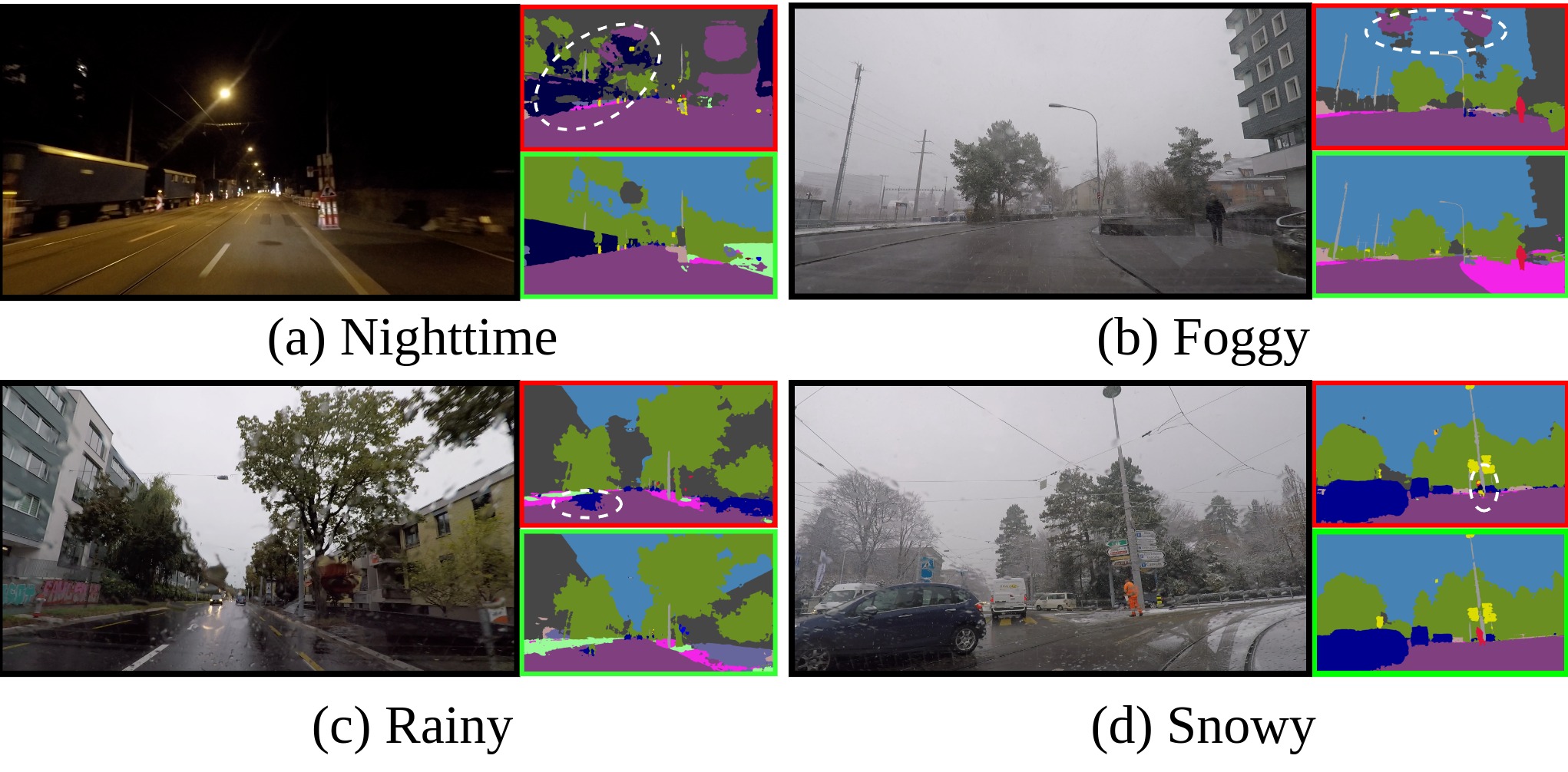}
    \caption{\textbf{Visualization of domain-generalized semantic segmentation results: MIC\cite{hoyer2023mic} (red box) vs. WeatherDG (green box).} The tested images include foggy, nighttime, snowy, and rainy scenarios.}
    \label{fig:vis_seg}
    \vspace{-6mm}
\end{figure}

\begin{figure}[h!]
  \centering
    \includegraphics[width=0.95\linewidth]{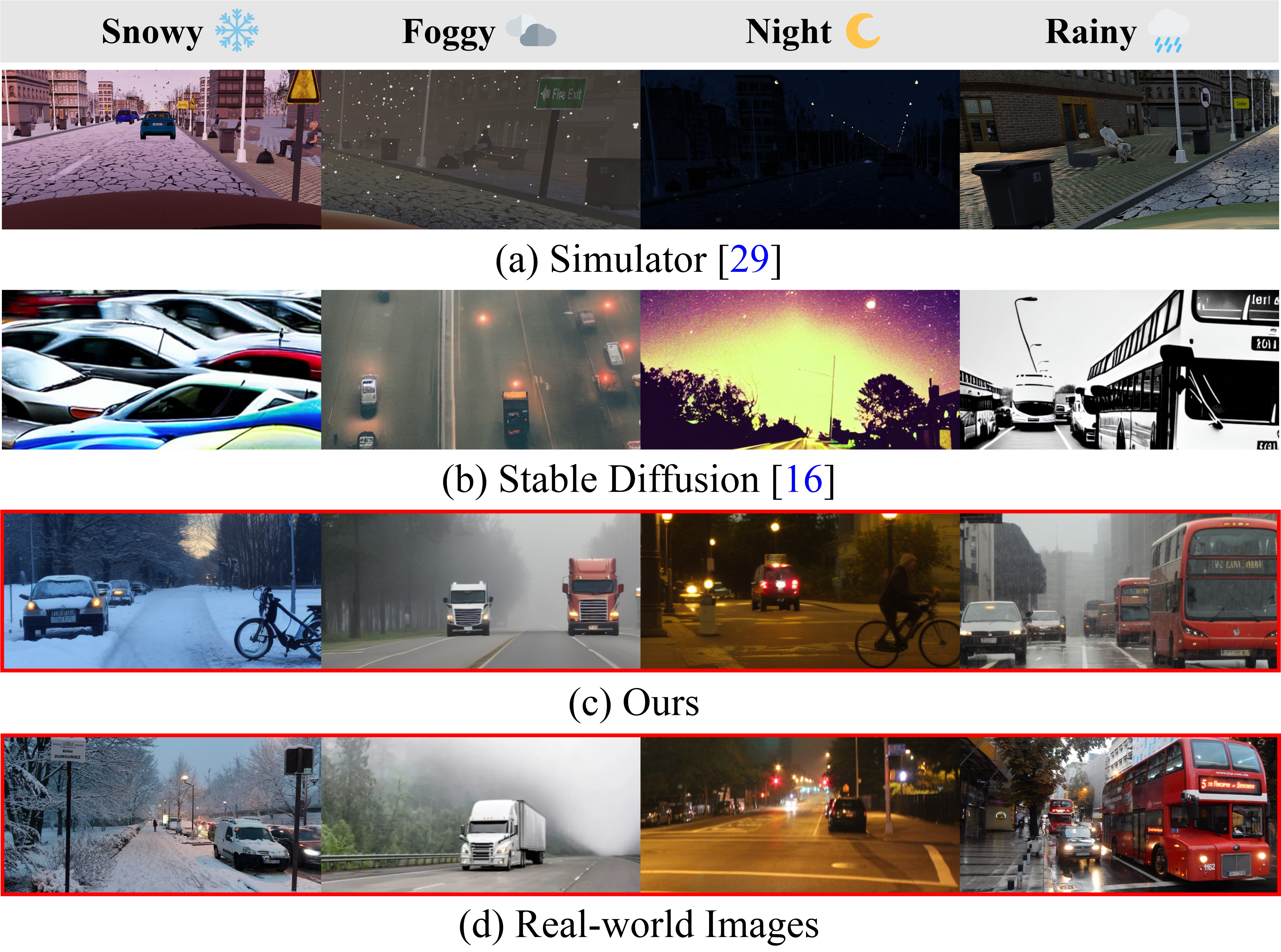}
  \caption{\textbf{Comparison between synthetic and real-world images under adverse weather conditions.} The results reveal that images generated by (a) driving simulator \cite{kerim2022Semantic} lack intricate details and natural lighting, whereas (b) Stable Diffusion \cite{rombach2021highresolution} typically produces images with an artistic flair. In contrast, (c) our method produces the most realistic images, closely resembling (d) diverse real-world driving scenes.} 
\vspace{-4mm}

\label{fig:Comp}
\end{figure}

Data augmentation for domain generalization aims to generate new, diverse and realistic synthetic images for augmenting the training data. Previous methods commonly adopt simulators \cite{Dosovitskiy17,Peng2017SimtoRealTO,9816133} or image translation models \cite{CycleGAN2017,huang2018munit} to generate new samples. Despite their effectiveness, these methods still suffer from diversity and authenticity, particularly in generating samples with adverse weather conditions (shown in \Cref{fig:Comp}). In recent years, Stable Diffusion (SD) \cite{rombach2021highresolution} has shown its strong ability to generate realistic, diverse, and high-quality images. This inspires us to leverage SD to solve the drawbacks of previous data augmentation methods in domain generalization. However, directly applying SD in our task will lead to a critical issue: the styles and layouts are very different from the driving-screen samples (see \Cref{fig:Comp}\textcolor{blue}{b})). Hence, training with such synthetic samples will hamper the performance of the model on unseen domains. This problem is mainly caused by that the SD was trained with various types of samples, such as natural images and artistic images, instead of specifically with the driving-screen samples. As such, the SD cannot well generate on-the-hand driving-screen-aware samples without a detailed and specific guide.

To solve the above drawback, we propose a novel data augmentation approach named WeatherDG to generate realistic, weather-diverse, and driving-screen images based on Stable Diffusion (SD) and Large Language Model (LLM). Our method is composed of three steps. \textit{Step-I: SD Fine-tuning}. We fine-tune the SD with source data. This enables us to align the content and layout of generated samples with real-world driving scenarios. \textit{Step-II: Procedural Prompt Generation}. We propose a prompt generation method based on LLM, where we leverage the LLM agents to procedurally enrich scenario descriptions (prompt). The generated prompt can help SD to automatically generate more diverse, detailed images. In addition, we introduce a balanced generation strategy to enrich tailed classes. \textit{Step-III: Sample Generation and model training.} Given the fine-tuned SD and the generated prompts, we can then generate new, diverse samples for model training. The generated samples are used to train the model together with the source data. In sum, our contributions are threefold:

\begin{itemize}
    \item We propose a novel data augmentation framework based on SD and LLM for domain generalization in adverse weather conditions. Our method can generate realistic, diverse samples for improving the model's generalization ability in unseen domains.

    \item We propose two novel strategies for prompt generation and sample generation, which encourage SD to generate diverse and driving-screen samples that are beneficial to our segmentation task.

    \item Our method is segmentation-model-agnostic. Experiments on three challenging datasets demonstrate that our method can consistently improve the performance of state-of-the-art methods.
\end{itemize}

\section{Related work}
\label{sec:realted_work}

\noindent \textbf{Domain  Generalization for Semantic Segmentation (DGSS)}. DGSS aims to train deep neural networks that perform well on semantic segmentation tasks across multiple unseen domains. Existing DGSS methods \cite{mancini2018robust,10083238,jiang2023domain} attempt to address the domain gap problem through two main approaches: normalization and data augmentation. Normalization-based methods \cite{peng2022semantic,choi2021robustnet} train by normalizing the mean and standard deviation of source features or whitening the covariance of these features. Data augmentation-based method \cite{jiang2023domain} transform source images into randomly stylized versions, guiding the model to capture domain-invariant shape features as texture cues are replaced with random styles \cite{pmlr-v205-niemeijer23a}. For instance, SHADE \cite{zhao2022shade} creates new styles derived from the foundational styles of the source domain, while MoDify \cite{jiang2023domain} utilizes difficulty-aware photometric augmentation.

With the advent of generative models, several studies \cite{Niemeijer2023DIDEX,Benigmim_2024_CVPR, qian2025allweather} have proposed input enhancement and content augmentation to enhance generalization. However, these approaches often either rely on weather normalization or lack realism in capturing variations in weather and lighting conditions.

\noindent \textbf{Unsupervised Domain Adaptation (UDA).}  UDA aims to boosts model performance on domain-specific data without needing labeled examples. Existing UDA techniques can be categorized into three main approaches: discrepancy minimization, adversarial training, and self-training. Discrepancy minimization reduces the differences between domains by using statistical distance functions \cite{sun2022safe}. Adversarial training involves a domain discriminator within a GAN framework to encourage domain-invariant input \cite{8460982}, feature \cite{diaz2020instance} or output \cite{tsai2018learning}. Self-training generates pseudo-labels for the target domain based on predictions made using confidence thresholds \cite{Zhang_2018_CVPR} or pseudo-label prototypes \cite{zhang2019category}. 
Recently, DATUM \cite{benigmim2023one} introduces a one-shot domain adaptation method that generates a dataset using a single image from the target domain and pairs it with unsupervised domain adaptation training methods to bridge the sim-to-real gap. Futher, PODA \cite{fahes2023poda} leveraged the capabilities of the CLIP model to enable zero-shot domain adaptation using prompts.

\noindent \textbf{Text-based Image Synthesis.} The current text-to-image task is predominantly driven by diffusion-based and LLM-oriented methods. Diffusion models, a breakthrough in producing photorealistic images, prompting studies to explore their use in enriching source domain datasets and improving semantic segmentation. For example, DIDEX \cite{Niemeijer2023DIDEX} employs ControlNet \cite{zhang2023adding} to convert synthetic images into real-world styles. Nevertheless, this method often lacks realism and replicates the spatial layout of the training data, limiting the diversity.

On the other hand, large language models also play a crucial role. CuPL \cite{pratt2023does} leverages GPT-3 \cite{brown2020language} to generate text descriptions that enhancing zero-shot image classification. CLOUDS \cite{Benigmim_2024_CVPR} uses Llama \cite{touvron2023llama} to create prompts for diffusion models. 

However, they fail to adequately consider the complexities introduced by varying weather and lighting conditions. In contrast, our approach employs a chain of LLMs acting as agents to not only craft detailed descriptions of complex real-world scenarios but also implement a tailored generation strategy. This ensures that the generated images are both diverse and realistic, and address the class imbalance problem in challenging conditions.


\begin{figure*}[h!]
  \centering
    \includegraphics[width=0.98\linewidth]{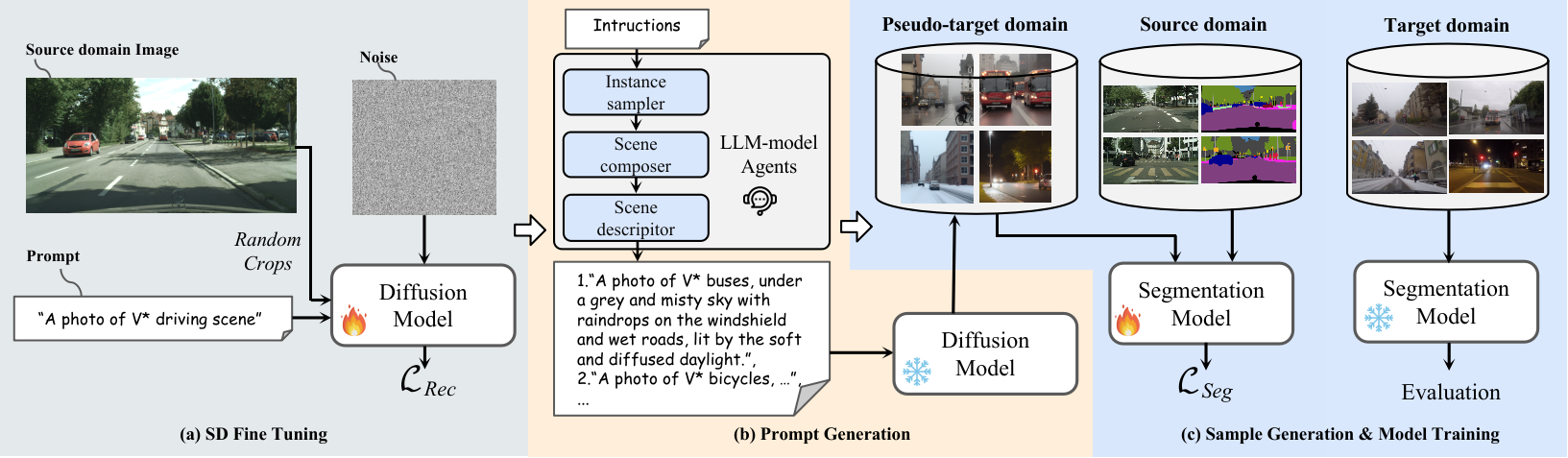}
  \caption{\textbf{The overview of WeatherDG pipeline.} (a) We first fine-tune a text-to-image diffusion model to integrate scene priors from the source domain. This ensures the images generated by the diffusion model accurately depict driving scenes. (b) Next, we employ a chain of LLM agents to procedurally construct detailed prompts that can enrich tailed classes and generate diverse weather and lighting effects with the fine-tuned model. (c) After generating images with these prompts, we utilize UDA techniques to train these images with the source domain dataset, followed by evaluation on real-world target datasets.}
    \label{fig:archi_overview}
\vspace{-3mm}
\end{figure*}

\section{Proposed Method}

\subsection{Overview}

\sysname~aims to generate images tailored to weather-specific autonomous driving scenes, enhancing semantic segmentation in challenging conditions. We begin by fine-tuning a diffusion model to adapt scene priors from the source domain, ensuring the generated images are within a driving scene (\Cref{sec:priors}). Next, we employ a procedural prompt generation method to create detailed prompts that enable the diffusion model to produce realistic and diverse weather and lighting effects (\Cref{sec:Procedural}). Additionally, we incorporate a probability-oriented sampling strategy for prompt generation. Following this, we use UDA training methods to leverage these generated images for semantic segmentation training (\Cref{sec:Domain}). The overview of the proposed approach is shown in \Cref{fig:archi_overview}. 

\subsection{SD Fine-tuning}
\label{sec:priors}
\begin{figure}[htb]
\vspace{-3mm}

  \centering
  \begin{subfigure}{\linewidth}  
    \centering
    \includegraphics[width=\linewidth, height=2.8in, keepaspectratio]{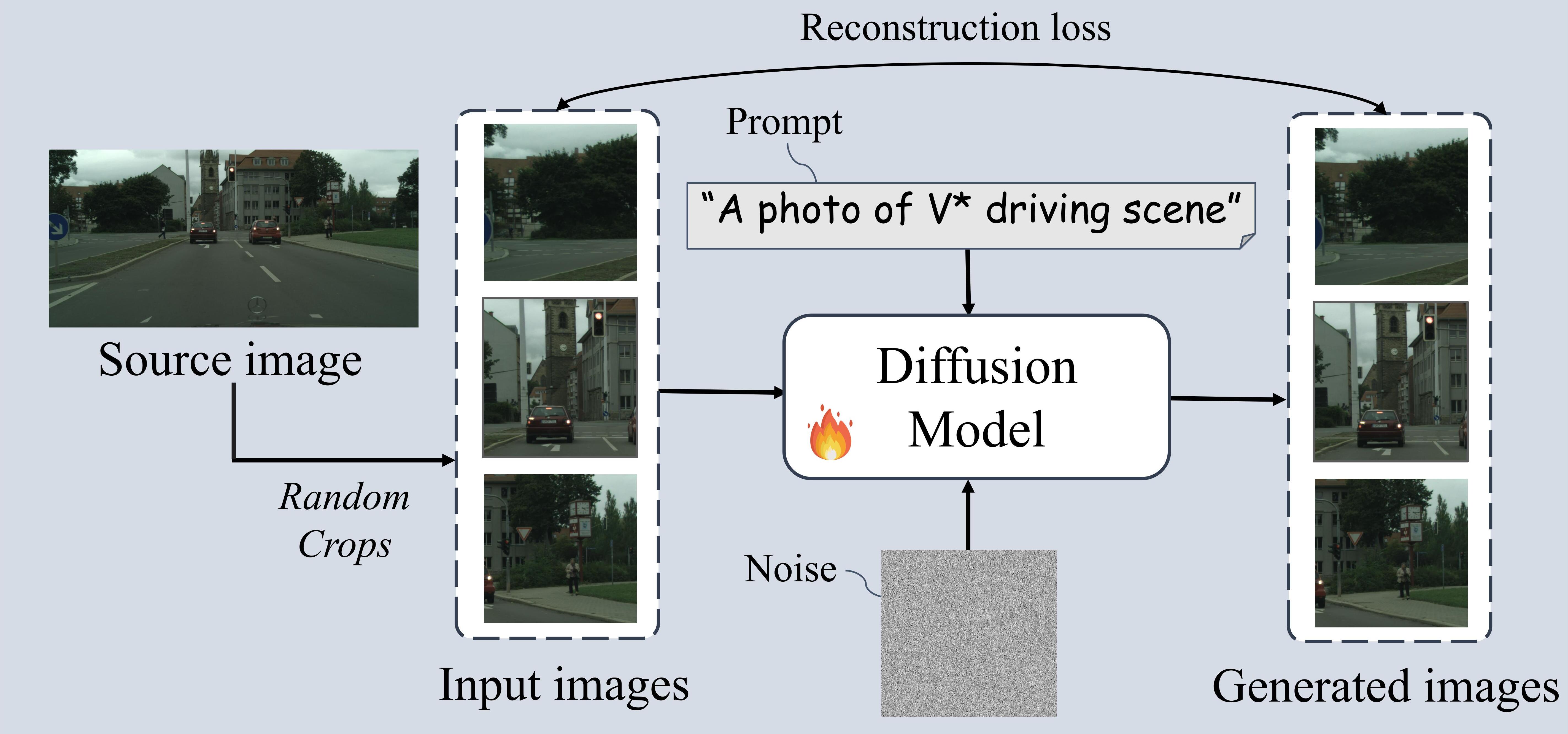}
    \subcaption{Training stage: A unique identifier, $V^{*}$, links prompts to specific image patches from the dataset, following related work to maintain language-scene association and prevent drift.}
    \label{fig:method_2}
  \end{subfigure}

  \begin{subfigure}{\linewidth}  
    \centering
    \includegraphics[width=\linewidth]{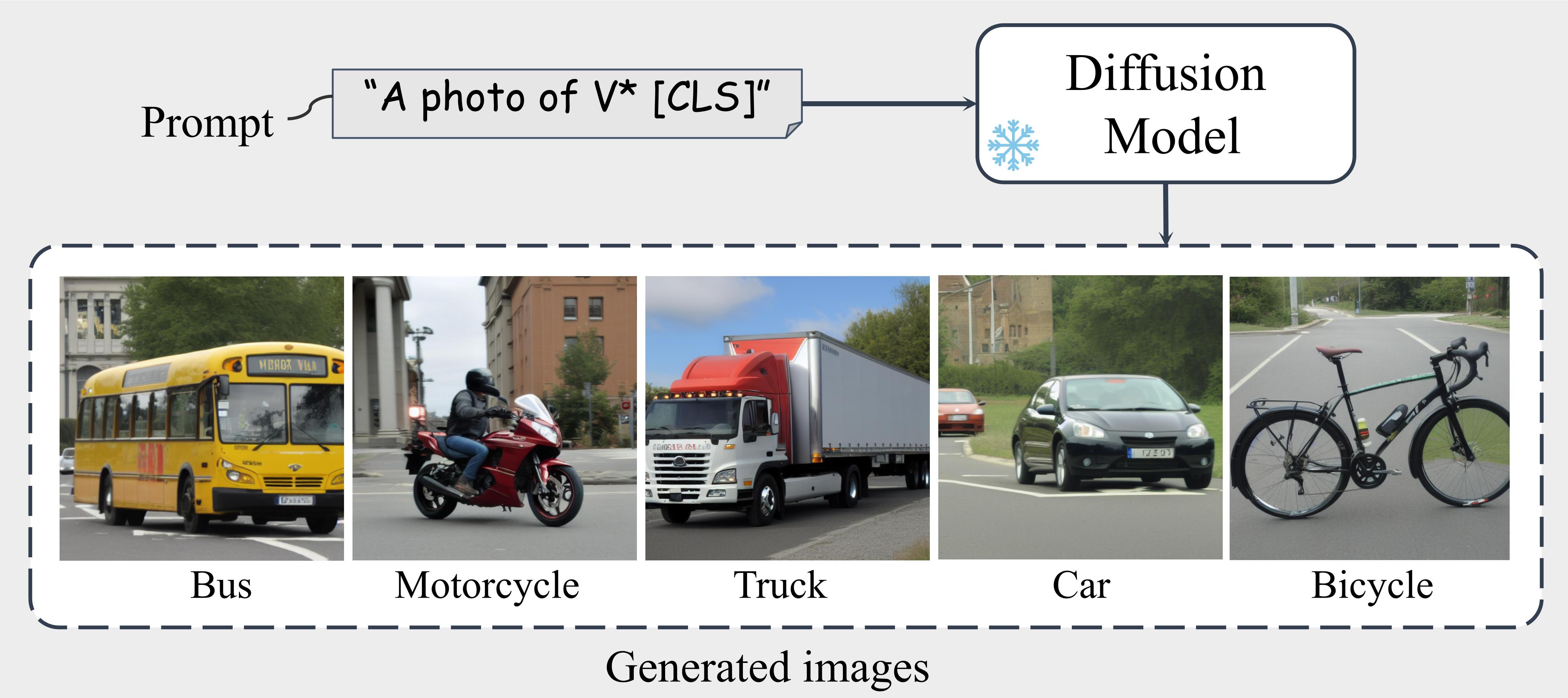}
    \subcaption{Inference stage: After training, the model uses the identifier $V^{*}$ to generate new instances in autonomous driving contexts, ensuring coherent scene synthesis.}
    \label{fig:method_3}
  \end{subfigure}
  \caption{The detailed process of Stable Diffusion \cite{rombach2021highresolution} fine-tuning.}
  \label{fig:scene_infer}
  
    \vspace{-6mm}

\end{figure}

Recently, diffusion models have advanced the field of generative domain adaptation, showcasing exceptional capabilities in producing photo-realistic images conditioned on text \cite{zhang2023adding}. However, directly applying diffusion models in autonomous driving setting presents challenges due to shifts in scene priors like style and layout. For example, the model often generates artistic images or adopt a bird's eye view perspective, which is different from the images in real-world autonomous driving datasets, as shown in Figure~\ref{fig:method_3}. When these images are incorporated during training, they can disrupt the knowledge the model has acquired from the labeled source domain, thereby harming the semantic segmentation performance.
To address these issues, we finetune a diffusion model \cite{rombach2021highresolution} to generate diverse images that retain content and layouts relevant to source domain. The input consists of a clean image paired with a corresponding prompt text, while the output is an image tailored to the autonomous driving domain.



We follow similar text-to-image diffusion model training procedures described in the relevant work \cite{ruiz2023dreambooth,benigmim2023one}. Specifically, we use a unique identifier in the prompt to link priors to one single image choosen from the source domain dataset. For instance, the prompt ``A photo of $V^{*}$ driving scene" associates with image patches cropped from the selected image, where $V^{*}$ identifies the scene, and ``driving scene" describes it broadly. This prevents language drift and improves model performance \cite{ruiz2023dreambooth}. The training procedure is presented in \Cref{fig:method_2}. After training, the model captures scene priors through the unique identifier $V_{*}$ and can generate new instances in similar contexts within autonomous driving datasets, as shown in Figure~\ref{fig:method_3}.

\subsection{Procedural Prompt Generation}
\label{sec:Procedural}
To enable the diffusion model to generate weather and lighting effects, it is essential to integrate specific weather conditions and times of day into the prompt. However, simplistic templates such as ``A photo of [CLS], [WEATHER], [TIME]'' often do not yield diversity and details, necessitating the need for more nuanced descriptions. Manually writing these descriptions is labor-intensive; therefore, we consider adopting LLM models to automate this process. Additionally, given the scarcity of dynamic object samples in adverse weather conditions, we need to consider a balanced generation strategy during the prompt generation to enrich these objects. To this end, our requirements for the generated prompts are threefold: 1) it should incorporate a balanced generation strategy, 2) introduce different weather and lighting conditions in an even manner, and 3) provide detailed descriptions of these conditions. Most importantly, all of the prompts are automatically generated by an LLM model to reduce human effort. During the implementation, we found that directly giving a single instruction for one LLM model fails to meet all of our requirements. Specifically, the generated prompts frequently do not align with our designed generation strategy or achieve the level of detailed description we expect. To address this, we develop a procedural prompt generation method involving a sequence of three LLM agents—namely, instance sampler, scene composer, and scene descriptor. This hierarchical approach enables us to generate precisely tailored text prompts for image generation, ensuring each aspect of the prompt aligns with our intended outcomes.



\paragraph{Instance sampler}
During the inference phase, we can generate a specific instance by simply adding the instance name \textless CLS \textgreater to the prompt, which serves as the task for the instance sampler. However, we notice that the semantic segmentation evaluation metrics for ``thing'' classes are significantly lower compared to ``stuff'' classes in most autonomous driving datasets. This disparity can be attributed to the class imbalance problem where ``stuff'' classes have more occurances in the images than ``thing'' classes. In particular, this issue is exacerbated under adverse weather conditions , as dynamic objects are less frequently present on the road in snowy or nightime scenes.

Therefore, we aim to enhance the instance sampler agent with the capability to employ a probability-oriented sampling strategy, giving underrepresented classes under adverse weathers higher sampling probabilities. This increases the likelihood of these rare classes presented in the generated images. In detail, we first compute the semantic label distribution of i-th thing classes $E_i$ in a typical adverse weather dataset \cite{sakaridis2021acdc} by: 
\begin{equation}
E_i = \frac{D_{\text{i}}}{D_{\text{thing}}},
\end{equation}
where $D_{\text{i}}$ represents number of semantics labels of each thing class,$D_{\text{ thing}}$ stands for number of all thing classes. Then the sampling probability $P_{\text{i}}$ can be formulated as:
\begin{equation}
P_i = \frac{1}{\sum_{j=1}^{n} E_j} \times \frac{1}{E_i}.
\end{equation}
Then, we allow the instance sampler agent to query the sampling probability to generate prompts. In this way, we can alleviate the shortage of instances that rarely appear in adverse weather conditions. 


\paragraph{Scene composer}

\begin{figure}[t!]
  \centering
  \begin{subfigure}{\linewidth}
    \centering
    \includegraphics[width=.95\linewidth]{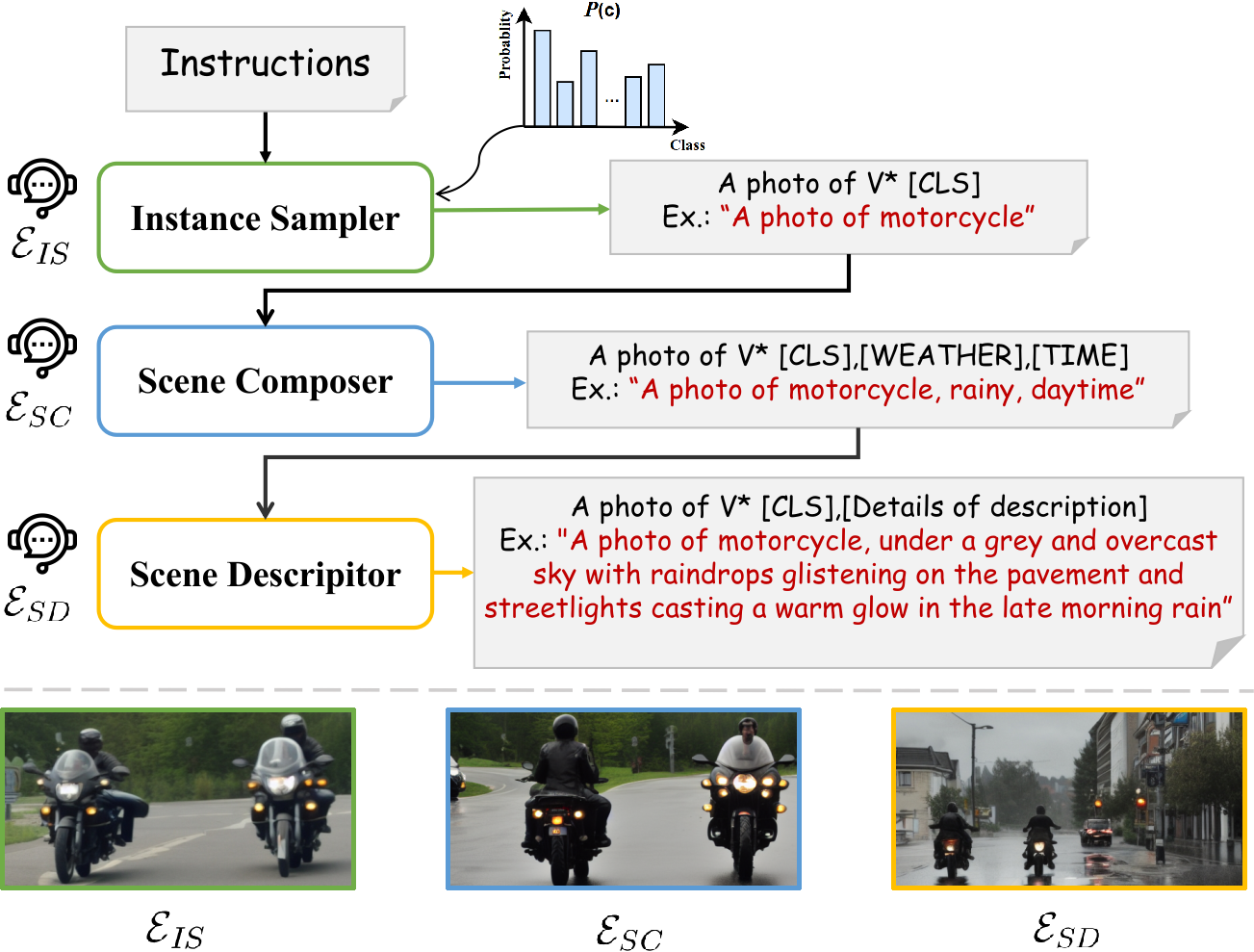}
  \end{subfigure}
  \caption{The process of prompt generation by gradually introducing LLM agents: \(\mathcal{E}_{\mathit{IS}}\), \(\mathcal{E}_{\mathit{SC}}\) and \(\mathcal{E}_{\mathit{SD}}\). The images correspond to the results generated using prompts created by different LLM agents.}
  \label{fig:llm-agent}
\end{figure}

Intuitively, to enable the generation of images covering a wide range of weather effects and different times of the day, we can design the prompt template as `A photo of \textless CLS \textgreater, \textless WEATHER \textgreater, \textless TIME \textgreater' to describe the image to be generated. Here, \textless CLS \textgreater\ refers to the classes in the typical autonomous driving dataset, \textless WEATHER \textgreater\ specifies the weather conditions, and \textless TIME \textgreater\ indicates the time of the day. To enhance the balance and diversity of the generated dataset, we include three common weather conditions: snowy, rainy, and foggy, along with two distinct times of day: daytime and nighttime. Each condition and time period is equally represented in the dataset to ensure comprehensive coverage and variability. 
However, merely incorporating categories of weather and time category does not guarantee the desired diversity in the effects, it often generates a singular subject with minimal environmental context and a limited range of effects in relatively simple scenes. As shown in Figure \ref{fig:llm-agent}, with the prompt ``A photo of motorcycle, rainy, daytime" generated by \(\mathcal{E}_{\mathit{SC}}\), the model adds reflections on the road in daytime lighting, indicating a humid, rainy day, but the effect is subtle and the scene details are limited. Although the result does not fully meet expectations, it provides a solid foundation for further improvement.


\paragraph{Scene descripitor}

To ensure the generated images reflect a variety of environmental effects and capture intricate details that resemble real-world scenes, we found that providing detailed descriptions of weather conditions and lighting is useful. For example, for the prompt ``A photo of a motorcycle in rainy, daytime", we enhance the ``rainy" aspect by including details such as ``under a grey and overcast sky with raindrops on the pavement". Similarly, for ``daytime", we add specifics like ``streetlights casting a warm glow in the late morning rain". Using the crafted prompt, the model incorporates elements like buildings, streetlights, a reflective road, and a misty sky into the scene, significantly enhancing the complexity and realism of the generated images. This enables the model trained with these images to generalize better to real-world scenarios.\\

However, manually crafting these detailed descriptions can be labor-intensive. By using LLM, we can automatically generate these nuanced elements, making the creation of highly detailed scenes more efficient. Consequently, we incorporate another LLM agent that receives prompts produced by \(\mathcal{E}_{\mathit{SC}}\) and uses the provided information to generate various detailed descriptions of scenes. In this way, we can enable a diversity of weather and lighting effects generation. 

\paragraph{Procedural generation}
The procedural prompt generation pipeline is as follows: first, the instance sampler \(\mathcal{E}_{\mathit{IS}}\) queries the sampling probability to select the class to be generated. Next, based on the instance sampler's output, the scene composer \(\mathcal{E}_{\mathit{SC}}\) composes scenes in the prompt that include typical weather and times of the day. Finally, after receiving the prompt generated by \(\mathcal{E}_{\mathit{SC}}\), the scene descriptor \(\mathcal{E}_{\mathit{SD}}\) creates detailed descriptions of the corresponding weather and lighting conditions. Taking the probability of the instance sampling as $\mathcal{P}$ , the process of generating text prompt $\mathcal{T}$ can be formulated as:
\[
\mathcal{T} = \mathcal{E}_{\mathit{SD}}\left(\mathcal{E}_{\mathit{SC}}\left(\mathcal{E}_{\mathit{IS}}(\mathcal{P})\right)\right).
\].


\vspace{-9mm}

\subsection{Sample Generation \& Model training}
\label{sec:Domain}
With the prompt $\mathcal{T}$, we send it to the fine-tuned diffusion model to generate image samples. Although the model can produce realistic images featuring challenging weather and lighting conditions, it is still difficult to incorporate pixel-level information because the images lack semantic labels.
To overcome this, we employ UDA methods such as DAFormer \cite{hoyer2022daformer} and HRDA \cite{hoyer2022hrda}, which facilitate the adaptation of a segmentation model to an unlabeled target dataset. Specifically, we train these methods using Cityscapes\cite{cordts2016cityscapes} dataset as the source domain, and our generated dataset as the pseudo-target domain. The model is then evaluated on the target three real-world benchmarks \cite{sakaridis2021acdc,yu2020bdd100k,SDV20}. The proposed framework can be adapted to UDA method easily, transforming it into a domain generalization method.


\section{Experiments}

\subsection{Dataset}
We conduct the experiments with domain generalization settings, using Cityscapes~\cite{cordts2016cityscapes} as source domain dataset, ACDC~\cite{sakaridis2021acdc}, BDD100k~\cite{yu2020bdd100k} and DarkZurich~\cite{SDV20} as test target domain dataset.  The Cityscapes dataset comprises 2,975 images captured under standard weather conditions and during daytime, each accompanied by corresponding semantic segmentation labels for training. The ACDC dataset features images taken under various typical weather and lighting conditions, including snowy, rainy, foggy, and nighttime settings. BDD100k contains various weathers and geographic locations. DarkZurich consists of images captured at night, providing a challenging environment for nighttime visual perception tasks. For fine-tuning the diffusion model, we use a single image sampled from the Cityscapes dataset. For unsupervised domain adaptation (UDA) training, we utilize the Cityscapes training set along with images generated by our diffusion model. For evaluation, we use 406 images from the ACDC validation set, 1,000 images from the BDD100k validation set, and 50 images from the DarkZurich validation set.

\subsection{Implementation details}
We utilize the Stable Diffusion \cite{rombach2021highresolution} as pretrained model and fine-tune it using DreamBooth \cite{ruiz2023dreambooth} method. During fine-tuning stage, we randomly crop a patch of 512$\times$512 size from the images selected from Cityscapes dataset. Then we pair it with customized prompt ``a photo of V driving scene". We use mean square error loss to quantify the differences between the original image patch and the generated image during model training. After training, we use text prompts generated by a chain of Llama \cite{touvron2023llama} models to create images. The generated images are further used in UDA training for generalizing normal weather source domain to adverse conditions domain.

\section{Evaluation and Discussion}

\subsection{Comparison with State-of-the-art }
We evaluate WeatherDG against state-of-the-art domain generalization models using ResNet-50 \cite{he2016deep} and MiT-B5 \cite{xie2021segformer} as encoders. As shown in \Cref{tab:performance}, for models using ResNet-50 as encoder, our model consistently outperforms state-of-the-art methods with the highest average mIoU score. With MiT-B5 as backbone, our model exceeds the second-best model MIC\cite{hoyer2023mic} on the ACDC and DarkZurich datasets by over 10\% and on the BDD100K dataset by 4.3\% in mIoU performance, achieving the best semantic segmentation performance. In addition, we visualize our model's semantic segmentation results under challenging conditions and compare them with MIC. As shown in \Cref{fig:vis_seg}, our model correctly segments sidewalks and sky in foggy and nighttime scenes. In the rainy scene, reflections on the road that are falsely recognized as vehicles by MIC are alleviated by our model. In the snowy scenario, our model successfully detects pedestrians on the road that MIC fails to recognize. These findings indicate our model's superior generalization capabilities compared to state-of-the-art methods in real-world challenging weather and lighting conditions. 
\begin{table}[h]
    \centering
    \resizebox{\linewidth}{!}{%
    \begin{tabular}{lccccc}
        \toprule
        \multirow{2}{*}{\textbf{Method}} & \multirow{2}{*}{\textbf{Encoder}} & \multicolumn{3}{c}{\textbf{Test domains mIoU}} & \multirow{2}{*}{\textbf{Avg.}} \\
        \cmidrule(lr){3-5}
        & & \textbf{ACDC} & \textbf{BDD100K} & \textbf{DarkZurich} & \\
        \midrule
        Source-only & \multirow{6}{*}{ResNet-50} & 35.9 & 37.3 & 9.0 & 27.4 \\
        IBN-Net\cite{pan2018IBN-Net} & & 42.0 & 45.8 & 17.3 & 35.0 \\
        RobustNet\cite{choi2021robustnet} & & 41.7 & 43.4 & 19.4 & 34.8 \\
        SHADE\cite{zhao2022shade} & & 42.1  & \textbf{49.1} & 22.6 & 37.9 \\
        DPCL\cite{DPCL} & & 43.8  & 44.9 & 23.4 & 37.3 \\

        \textbf{WeatherDG (ours)} & & \textbf{45.2} & 45.8 & \textbf{23.5} & \textbf{38.2} \\
        \midrule
        Source-only & \multirow{5}{*}{MiT-B5} & 44.6  & 44.7 & 18.7 & 36.0 \\
        DAFormer\cite{hoyer2022daformer} & & 47.8 & 49.2 & 24.7 & 40.6 \\
        HRDA\cite{hoyer2022hrda} & & 46.3 & 49.8 & 25.8 & 40.6 \\
        MIC\cite{hoyer2023mic} & & 50.0 & 53.1  & 21.5 & 41.5 \\
        \textbf{WeatherDG (ours)} & & \textbf{60.2} & \textbf{57.4}  & \textbf{35.3} & \textbf{51.0} \\
        \bottomrule
    \end{tabular}
    }
        \caption{\textbf{Domain generalization performance (mIoU (\%)) of state-of-arts methods using ResNet-50 and MiT-B5 as encoders}. The compared methods are retrained with Cityscapes dataset. Evaluations are performed on ACDC, BDD100K, and DarkZurich datasets that feature adverse conditions, such as snow, rain, fog, and low-light scenarios.}
    \label{tab:performance}
    \vspace{-6mm}
\end{table}

\begin{figure*}[htb]
        \includegraphics[width=\linewidth, height=2.8in, keepaspectratio]{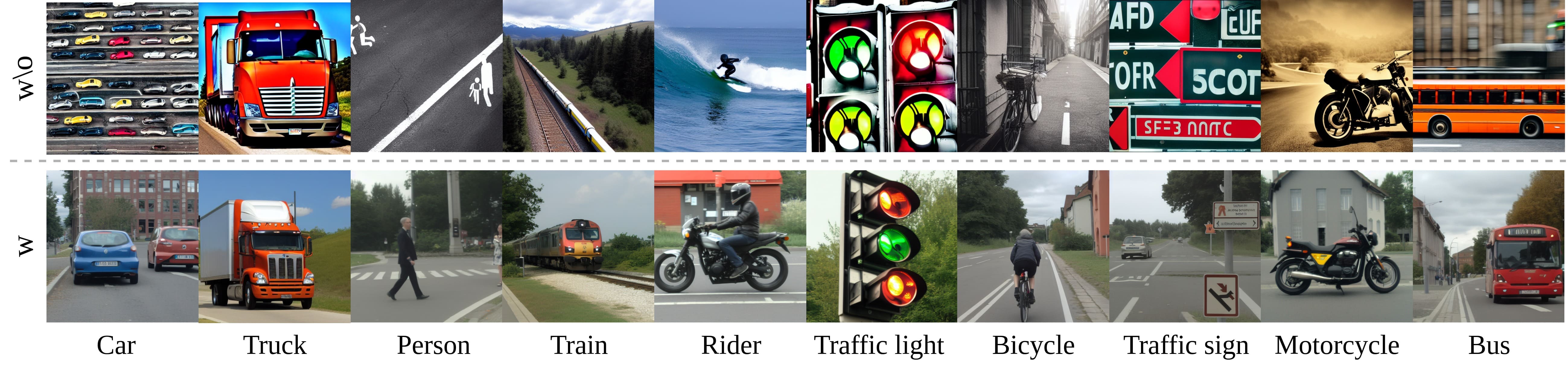}
        \caption{\textbf{Comparison of images generated by the plain stable diffusion model (top row) and our fine-tuned model (bottom row) using the same prompt template}. The results illustrate that our fine-tuned model significantly reduces artistic and unrealistic elements, generating images more aligned with real-world autonomous driving scenarios.}
        \label{fig:scene_prior_results}
        \vspace{-3mm}

\end{figure*}
\subsection{Influence of UDA methods for training}
\begin{table}[htb]
\centering
\resizebox{\linewidth}{!}{
\begin{tabular}{lcccccc}
\toprule
\multirow{2}{*}{\textbf{Method}} & \multirow{2}{*}{\textbf{Encoder}} & \multicolumn{3}{c}{\textbf{Test domains mIoU}} & \multirow{2}{*}{\textbf{Avg.}} \\
\cmidrule(lr){3-5}
 & & \textbf{ACDC} & \textbf{BDD100K} & \textbf{DarkZurich} & \\
\midrule
DAFormer\cite{hoyer2022daformer} & \multirow{3}{*}{ResNet-50} & \textbf{45.2} & \textbf{45.8} & \textbf{23.5} & \textbf{38.2} \\
MIC\cite{hoyer2023mic} &  & 43.8 & 43.7 & 22.6 & 36.7 \\
HRDA\cite{hoyer2022hrda} &  & 44.5 & 44.6 & 22.4 & 37.2 \\
\midrule
DAFormer\cite{hoyer2022daformer} & \multirow{3}{*}{MiT-B5} & 53.3 & 53.5 & 24.7 & 43.8 \\
MIC\cite{hoyer2023mic} &  & 60.0 & 54.4 & 32.6 & 49.0 \\
HRDA\cite{hoyer2022hrda} &  & \textbf{60.2} &\textbf{ 57.4} & \textbf{35.3} & \textbf{51.0} \\
\bottomrule
\end{tabular}}
\caption{\textbf{Comparison of mIoU performance of UDA methods} trained using the labeled Cityscapes dataset as the source dataset and our generated unlabeled images as the target dataset.}
\label{tab:UDA}
\vspace{-3mm}
\end{table}
We investigate the influence of UDA methods by utilizing three different state-of-the-art approaches including DAFormer \cite{hoyer2022daformer}, HRDA \cite{hoyer2022hrda}, and MIC \cite{hoyer2023mic}, each trained with Cityscapes and our generated dataset. As shown in \Cref{tab:UDA}, DAFormer demonstrates superior adaptation to the pseudo-target domain among the ResNet-50 models, while HRDA achieves the best generalization performance across the three domains with the MiT-B5 encoder. To further study the influence of these methods under different conditions, we evaluate them across four typical challenging scenarios in the ACDC dataset. As indicated in \ref{tab:weather_miou}, all methods perform best in foggy conditions and worst in nighttime conditions. Notably, none of these methods exceed a 40\% performance in nighttime scenes, which is significantly lower than in other scenarios. This could be due to the substantial appearance differences between nighttime scenes and the predominantly daytime images in the training source domain, making adaptation challenging. This finding suggests that simply adding a pseudo-target dataset for adaptive training may be inadequate for complete knowledge transfer in the nighttime domain, necessitating more advanced adaptation techniques.
\begin{table}[htb]
\centering
\resizebox{\linewidth}{!}{
\begin{tabular}{lccccc}
\toprule
\multirow{2}{*}{\textbf{Method}} & \multicolumn{4}{c}{\textbf{Test weathers}} & \multirow{2}{*}{\textbf{Avg.}} \\
\cmidrule(lr){2-5}
 & \textbf{Snow} & \textbf{Foggy} & \textbf{Rainy} & \textbf{Nighttime} & \\
\midrule
Source-only & 49.9 & 61.9 & 47.8 & 19.6 & 44.8 \\
DAFormer\cite{hoyer2022daformer} & 54.2 & 66.8 & 54.1 & 27.3 & 50.6 \\
MIC\cite{hoyer2023mic} & 59.4 & 74.2 & 62.0 & 36.9 & 58.0 \\
HRDA\cite{hoyer2022hrda} & \textbf{59.7} & \textbf{75.9} & \textbf{64.6} & \textbf{38.7} & \textbf{59.7} \\
\bottomrule
\end{tabular}}
\caption{Comparison of mIoU performance of UDA techniques across typical weather and lighting conditions.}
\label{tab:weather_miou}
\vspace{-3mm}
\end{table}

\subsection{Influence of SD Fine-tuning.}
To demonstrate the influence of scene prior adaptation, we compare images generated by the plain stable diffusion model and our fine-tuned model. We use the same prompt, templated as ``A photo of [CLS]" for each model to generate commonly seen objects in the autonomous driving dataset.
\begin{table}[htb]
\centering
\label{tab:tab_scene}

\begin{tabular}{@{}lccc@{}} 
\toprule
\textbf{Method} & \textbf{ACDC} & \textbf{BDD100K} & \textbf{DarkZurich} \\
\midrule
w/o Fine-tuning & 49.9 & 52.1 &  24.7 \\
w Fine-tuning    & \textbf{50.8} & \textbf{53.2} &  \textbf{25.6}\\
\bottomrule

\end{tabular}
\caption{Comparison of mIoU performance of models trained on datasets generated with and without fine-tuned model.}
\label{tab:sd}
\vspace{-2mm}
\end{table}
The results in \Cref{fig:scene_prior_results} show that the plain stable diffusion tends to generate images with an artistic style or cinematic photography effects, as seen with ``truck," ``bicycle," ``motorcycle," and ``bus". For ``car" and ``train", the images present different camera perspectives, such as bird's-eye view. Additionally, for ``traffic light", ``traffic sign", and ``person", the model exhibits excessive creativity, generating overly stylized traffic lights and rendering ``person" as sketches. For the ``rider" category, the stable diffusion model imagines a surfer on the sea.

These samples may acquire incorrect pseudo-labels that negatively impact adaptive training and harm segmentation performance. In contrast, our fine-tuned model ensures that the generated instances are contextually appropriate for the driving scene, resulting in more accurate pseudo-labels and better adaptation to the autonomous driving domain. As shown in \Cref{tab:sd}, the performance of semantic segmentation model trained with images generated by the fine-tuned model is better than plain stable diffusion.


\begin{figure*}[htb]
    \centering
    \includegraphics[width=\linewidth]{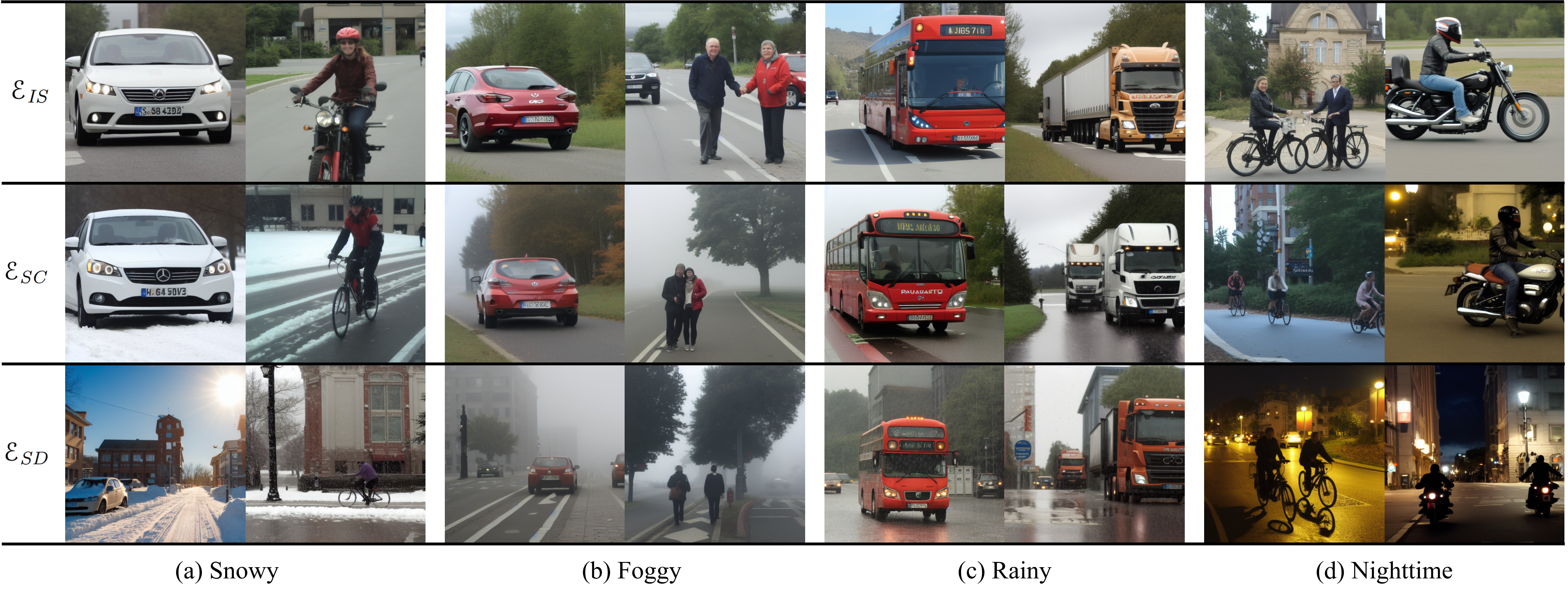}
    \caption{\textbf{Comparison of images generated using prompts created by different LLM agents.} Each row represents images generated by a specific LLM agent, while each column showcases images generated by different LLM agents for specific weather or lighting effects during the procedural prompt generation. The results demonstrate that \(\mathcal{E}_{\mathit{IS}}\) (top row) enables the model to generate diverse instances, albeit with limited scene detail. While \(\mathcal{E}_{\mathit{SC}}\) (middle row) allows the model to generate weather and lighting effects, the overall impact is rather subtle. With detailed descriptions crafted by \(\mathcal{E}_{\mathit{SD}}\), the model (bottom row) produces intricate scene details and more diverse weather and lighting effects, significantly enhancing the variety and realism of the generated images.}
    \label{fig:llm_res}
    \vspace{-3mm}
\end{figure*}
\subsection{Influence of Procedural Prompt Generation}
To illustrate the effectiveness of our procedural prompt generation, we compare images generated by text prompts created by different LLM agents in \Cref{fig:llm_res}. The results show that the instance sampler can generate the desired objects with a basic prompt such as ``A photo of [CLS]". By specifying the general category of weather and time of day, the scene composer only adds basic weather effects to the image, though these effects are subtle. Additionally, the generated samples primarily focus on the subject, often lacking scene details. When the scene descriptor crafts more detailed scene descriptions in the prompt, the model produces images with various realistic weather and lighting effects. As shown in third row, for snowy weather, the model generates a complex scene with heavy accumulated snow on the road and even snowflakes in the air. For rainy weather, the prompt generates reflections, raindrops, and a misty effect, indicating heavy rain. For nighttime, we can see that \(\mathcal{E}_{\mathit{SC}}\) fails to add sufficient nighttime lighting, but with prompts generated by \(\mathcal{E}_{\mathit{SD}}\), the scene in the image exhibit a darker tone and more intricate details, creating a more realistic nighttime environment. In addition, we evaluate the mIoU performance of the DAFormer\cite{hoyer2022daformer} trained with datasets generated using different LLM agents. In \Cref{tab:agent}, the results demonstrate that progressively refining the base prompt ``A photo of [CLS]" using these LLM agents results in higher mIoU scores.

\begin{table}[ht]
\centering
\begin{tabular}{lcccc}
\toprule
\multirow{2}{*}{\textbf{Models}} & \multicolumn{3}{c}{\textbf{LLM-Agents}} & \multirow{2}{*}{\textbf{mIoU}} \\
\cmidrule(lr){2-4}
 & \(\mathcal{E}_{\mathit{IS}}\) & \(\mathcal{E}_{\mathit{SC}}\) & \(\mathcal{E}_{\mathit{SD}}\) & \\
\midrule
$M_{base}$ & -- & -- & -- & 50.8 \\
$M_1$ & \checkmark & -- & -- & 51.5 \\
$M_2$ & \checkmark & \checkmark & -- & 52.1\\
$M_3$ & \checkmark & \checkmark & \checkmark & 53.3 \\
\bottomrule
\end{tabular}
\caption{Comparison of mIoU performance for models trained on datasets generated by introducing different LLM agents.}
\label{tab:agent}
\vspace{-2mm}
\end{table}

\subsection{Influence of Numbers of Generated Images}
\begin{figure}[t!]
    \centering
    \includegraphics[width=\linewidth]{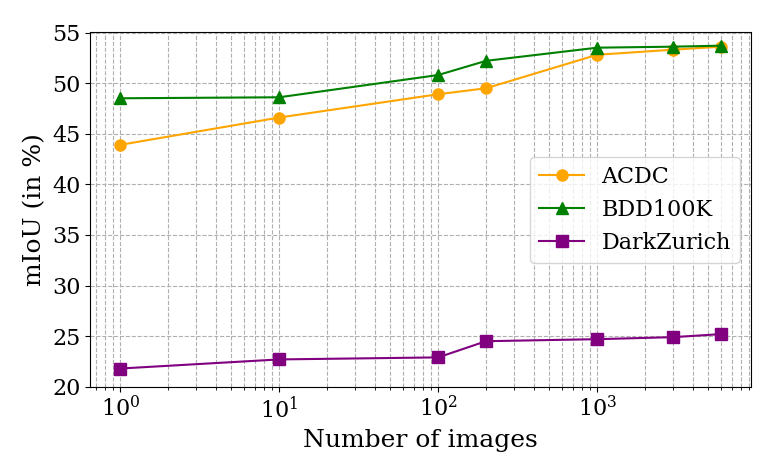}

    \caption{Impact of the number of generated images on training model's mIoU performance.}
    \label{fig:num_generation}
\vspace{-3mm}
\end{figure}

To investigate the impact of the generated dataset size on the performance of the segmentation model, we evaluate the mIoU performance of the DAFormer model trained with the generated dataset on the ACDC, BDD100K, and DarkZurich datasets. As shown in \Cref{fig:num_generation}, a substantial performance gain are observed as the number of images increases from 10 to 1000 for all three datasets, after which the performance gains level off. Notably, the Dark Zurich dataset shows much smaller improvements with the increased number of images compared to the other two datasets. This can be attributed to the inherent learning difficulty for nighttime images during the model training. These findings indicate that: 1) while increasing the amount of training data generally enhances model performance, the benefits may plateau after some point; and 2) inherent dataset characteristics can impact the extent of the model's improvement.

\section{Conclusion}

In this paper, we present WeatherDG, a novel approach for domain generalization in semantic segmentation under adverse weather conditions. By combining Stable Diffusion (SD) with a Large Language Model (LLM), our method enables automated generation of realistic images resembling real-world driving scenarios. Fine-tuning SD, along with procedural prompt generation and a balanced strategy, creates diverse weather effects and enhances tailed classes in generated images. These images, combined with source data, improve model generalization. Experiments across challenging datasets show that WeatherDG significantly boosts semantic segmentation performance, setting a new benchmark for robustness in autonomous driving.

{\scriptsize
\bibliographystyle{IEEEtran}
\bibliography{references}
}

\end{document}



\section{Introduction}

After receiving paper reviews, authors may optionally submit a rebuttal to address the reviewers' comments, which will be limited to a {\bf one page} PDF file.
Please follow the steps and style guidelines outlined below for submitting your author response.

The author rebuttal is optional and, following similar guidelines to previous conferences, is meant to provide you with an opportunity to rebut factual errors or to supply additional information requested by the reviewers.
It is NOT intended to add new contributions (theorems, algorithms, experiments) that were absent in the original submission and NOT specifically requested by the reviewers.
You may optionally add a figure, graph, or proof to your rebuttal to better illustrate your answer to the reviewers' comments.

Per a passed 2018 PAMI-TC motion, reviewers should refrain from requesting significant additional experiments for the rebuttal or penalize for lack of additional experiments.
Authors should refrain from including new experimental results in the rebuttal, especially when not specifically requested to do so by the reviewers.
Authors may include figures with illustrations or comparison tables of results reported in the submission/supplemental material or in other papers.

Just like the original submission, the rebuttal must maintain anonymity and cannot include external links that reveal the author identity or circumvent the length restriction.
The rebuttal must comply with this template (the use of sections is not required, though it is recommended to structure the rebuttal for ease of reading).


\subsection{Response length}
Author responses must be no longer than 1 page in length including any references and figures.
Overlength responses will simply not be reviewed.
This includes responses where the margins and formatting are deemed to have been significantly altered from those laid down by this style guide.
Note that this \LaTeX\ guide already sets figure captions and references in a smaller font.

\section{Formatting your Response}

{\bf Make sure to update the paper title and paper ID in the appropriate place in the tex file.}

All text must be in a two-column format.
The total allowable size of the text area is $6\frac78$ inches (17.46 cm) wide by $8\frac78$ inches (22.54 cm) high.
Columns are to be $3\frac14$ inches (8.25 cm) wide, with a $\frac{5}{16}$ inch (0.8 cm) space between them.
The top margin should begin 1 inch (2.54 cm) from the top edge of the page.
The bottom margin should be $1\frac{1}{8}$ inches (2.86 cm) from the bottom edge of the page for $8.5 \times 11$-inch paper;
for A4 paper, approximately $1\frac{5}{8}$ inches (4.13 cm) from the bottom edge of the page.

Please number any displayed equations.
It is important for readers to be able to refer to any particular equation.

Wherever Times is specified, Times Roman may also be used.
Main text should be in 10-point Times, single-spaced.
Section headings should be in 10 or 12 point Times.
All paragraphs should be indented 1 pica (approx.~$\frac{1}{6}$ inch or 0.422 cm).
Figure and table captions should be 9-point Roman type as in \cref{fig:onecol}.

List and number all bibliographical references in 9-point Times, single-spaced,
at the end of your response.
When referenced in the text, enclose the citation number in square brackets, for example~\cite{Alpher05}.
Where appropriate, include the name(s) of editors of referenced books.

\begin{figure}[t]
  \centering
  \fbox{\rule{0pt}{0.5in} \rule{0.9\linewidth}{0pt}}
   \caption{Example of caption.  It is set in Roman so that mathematics
   (always set in Roman: $B \sin A = A \sin B$) may be included without an
   ugly clash.}
   \label{fig:onecol}
\end{figure}

To avoid ambiguities, it is best if the numbering for equations, figures, tables, and references in the author response does not overlap with that in the main paper (the reviewer may wonder if you talk about \cref{fig:onecol} in the author response or in the paper).
See \LaTeX\ template for a workaround.

\subsection{Illustrations, graphs, and photographs}

All graphics should be centered.
Please ensure that any point you wish to make is resolvable in a printed copy of the response.
Resize fonts in figures to match the font in the body text, and choose line widths which render effectively in print.
Readers (and reviewers), even of an electronic copy, may choose to print your response in order to read it.
You cannot insist that they do otherwise, and therefore must not assume that they can zoom in to see tiny details on a graphic.

When placing figures in \LaTeX, it is almost always best to use \verb+\includegraphics+, and to specify the  figure width as a multiple of the line width as in the example below
{\small\begin{verbatim}
   \usepackage{graphicx} ...
   \includegraphics[width=0.8\linewidth]
                   {myfile.pdf}
\end{verbatim}
}

{
    \small
    \bibliographystyle{ieeenat_fullname}
    \bibliography{main}
}